%% file: main.tex
\documentclass[runningheads]{llncs}
\usepackage[T1]{fontenc}
\usepackage{graphicx}
\usepackage{amsmath}
\usepackage{float}    % pour [H]
\usepackage{placeins} % pour \FloatBarrier
\usepackage{multirow}
\usepackage{tikz}
\usepackage{amsfonts}
\usetikzlibrary{arrows.meta, positioning}
\usepackage{subcaption}
\usepackage{booktabs}
\usepackage{makecell}

\usepackage{adjustbox} 
\usepackage{enumitem} 
\usepackage{etoolbox}
\usepackage{xurl}
\usepackage{hyperref}
\begin{document}

\title{DocQT: Improving Document Forgery Localization Robustness via Diverse JPEG Quantization Tables}
\titlerunning{DocQT for Forgery Localization Robustness}
\author{Kylian Ronfleux--Corail\inst{1,2}\orcidID{0009-0006-6601-4212} \and Guillaume Bernard\inst{1}\orcidID{0000-0001-5945-4865} \and Mickaël Coustaty\inst{2}\orcidID{0000-0002-0123-439X} \and Nicolas Sidère\inst{2}\orcidID{0000-0001-6719-5007}}
\authorrunning{K. Ronfleux--Corail et al.}

\institute{MAIF, Niort, France \and L3i Laboratory, La Rochelle University, La Rochelle, France}

\maketitle

\input{sections/abstract}
\input{sections/introduction}

\input{sections/sota}
\input{sections/donnees}

\input{sections/benchmark_protocol}
\input{sections/Results}
\input{sections/Conclusion}

% \vspace{-.4cm}
\subsubsection{\ackname}
This research was supported by the CIFRE PhD program funded by the ANRT.
% \vspace{-.4cm}

\begin{credits}
  \subsubsection{\discintname} Kylian Ronfleux--Corail and Guillaume Bernard are employed by MAIF, which provided the operational corpus. Mickaël Coustaty and Nicolas Sidère declare no additional competing interests.
\end{credits}
% \vspace{-.3cm}

\bibliographystyle{splncs04}
\bibliography{biblio}
\end{document}

%% file: sections/abstract.tex
\begin{abstract}
  Document manipulation localization models achieve strong performance on public benchmarks yet fail to generalize to operational document workflows. We identify a critical and overlooked source of this gap: the mismatch between the narrow distribution of JPEG quantization tables used during training --- restricted to standard libjpeg quality factors --- and the heterogeneous compression profiles encountered in real-world insurance document pipelines. To isolate this factor, we conduct a controlled factorial study comparing two architectures with contrasting levels of quantization table awareness --- FFDN~\cite{chen_enhancing_2025} and Mesorch~\cite{zhu_mesoscopic_2024} --- each trained under either standard quality factor augmentation (\textit{Standard-QT}) or operationally calibrated quantization tables sampled from \textsc{DocQT}, a quantization-table bank derived from a MAIF operational image corpus (\textit{Real-QT}), and evaluated under three recompression conditions. Training under \textit{Real-QT} yields substantial localization gains on DocTamper~\cite{qu_towards_2023} and significantly reduces the pixel-level false positive rate on authentic operational documents, but only for architectures that explicitly ingest the quantization table as input. The released \textsc{DocQT} quantization-table dataset and compression-reproduction material are directly available at \url{https://github.com/Kyliroco/Improving-Document-Forgery-Localization-Robustness-via-Diverse-JPEG-Quantization-Tables}. These results demonstrate that standard quality factor augmentation does not adequately proxy operational compression diversity, and that architectural choices explicitly conditioning on the quantization table provide a meaningful robustness advantage for real-world deployment.
  \keywords{Document Manipulation Localization \and JPEG Quantization Tables \and Image Forensics \and Compression Robustness}
\end{abstract}

%% file: sections/introduction.tex
\section{Introduction}
\label{sec:introduction}
\subsection{Document Manipulation Localization in Insurance Operational Workflows}
\label{sec:context}

Organizations such as insurance companies, public agencies, and financial institutions handle thousands of documents every day---ranging from claims and contracts to invoices and supporting paperwork. These documents constitute a primary vector for fraud and money laundering attempts~\cite{artaud_find_2018,martinez_tornes_jeu_2023,dong_robust_2024}. As an illustrative example of operational scale, internal observations at MAIF, a French insurance company, indicate that over 100{,}000 documents are received per month for claim management alone. At such volumes, manual inspection is impractical, and automated localization systems are required~\cite{artaud_find_2018}. In this context, false positives---that is, authentic documents incorrectly flagged as tampered---trigger additional manual reviews, increasing both processing time and operational costs. As argued by Guillaro~et~al.~\cite{guillaro_trufor_2023}, in a realistic deployment where manipulated images are rare, a high false alarm rate can render a system counterproductive, with false positives drastically outnumbering true positives.

In this work, we focus on \textit{Document Manipulation Localization} (DML), defined as the task of producing a pixel-level segmentation mask identifying tampered regions, independently of any intent classification. Importantly, the presence of tampering does not imply fraudulent intent: many legitimate documents undergo modifications such as highlighting, digital annotations, or watermarking that do not constitute forgery~\cite{martinez_tornes_jeu_2023}. Fraud determination remains a business-level decision~\cite{artaud_find_2018}. Throughout this paper, the terms \textit{tampering} and \textit{manipulation} are used interchangeably to denote pixel-level alteration; \textit{forgery} is reserved for cases in which the intent to deceive is established.

The majority of manipulation localization models are developed and evaluated on natural images, while document-oriented methods remain comparatively scarce~\cite{du_forensichub_2025,qu_towards_2023}. We argue that the performance reported on public benchmarks such as DocTamper~\cite{qu_towards_2023} overstates the practical robustness of these models when deployed in operational workflows. This paper investigates the primary source of this generalization gap: the mismatch between the narrow distribution of JPEG quantization tables used during model training and the heterogeneous compression profiles present in real-world document pipelines.

\subsection{Problem Statement: Quantization Table Mismatch}
\label{sec:problem_statement}

JPEG compression artifacts constitute a central forensic signal exploited by state-of-the-art localization models~\cite{kwon_learning_2022,qu_towards_2023,chen_enhancing_2025}. However, the forensic cues these models learn---in particular double-quantization residuals observed in the DCT domain---depend directly on the specific identity of the quantization table rather than merely a nominal quality factor~\cite{wan_overview_2023,kwon_learning_2022}. Most existing training pipelines restrict JPEG augmentation to standard quality factors derived from \textit{libjpeg}, which correspond to a small, discrete subset of all possible quantization tables~\cite{wan_overview_2023,neal_krawetz_pictures_2007}.

This assumption breaks down in operational settings. A MAIF operational image corpus reveals a broad diversity of acquisition devices, scanning software, and document conversion chains encountered in insurance workflows~\cite{nayef_smartdoc-qa_2015}; from this corpus, we derive \textsc{DocQT}, a dataset of \textbf{859 distinct luminance quantization tables}. In comparison, public benchmarks such as DocTamper~\cite{qu_towards_2023} and T-SROIE~\cite{wang_tampered_2022} expose only a single quantization table, while SROIE~\cite{huang_icdar2019_2019} contains at most 6 distinct configurations. This mismatch between training-time and inference-time quantization table distributions is expected to degrade localization accuracy and increase false positives on intact documents~\cite{kwon_learning_2022,wan_overview_2023}.

To address this gap, we conduct a comparative experimental study that systematically controls the quantization table distribution during both training and evaluation, following the evaluation framework of ForensicHub~\cite{du_forensichub_2025} and the benchmarking methodology of IMDL-BenCo~\cite{ma_imdl-benco_2024}. Our contributions are twofold: first, we quantify the domain shift induced by the mismatch between training-time quantization tables and those encountered in operational workflows; second, we demonstrate that retraining with operational insurance quantization tables significantly reduces false positives on authentic documents while preserving localization performance on tampered ones.

%% file: sections/sota.tex
\section{Related Works}
\label{sec:related_works}

\subsection{Document Manipulation Localization}
\label{sec:dml_sota}

Document manipulation localization is an active research field in image manipulation detection and localization, first studied extensively on natural scene images before being adapted to the specific challenges of documents~\cite{du_forensichub_2025,ma_imdl-benco_2024}. In this section, we review both lines of work, focusing on methods related to the JPEG compression artifacts that are central to our problem.

\subsubsection{Alteration Detection and Localization in Natural Images.}
\label{sec:natural_localization}

Natural images constitute the primary domain in which manipulation localization methods have been developed~\cite{du_forensichub_2025,chen_enhancing_2025,liu_pscc-net_2022}. A recurring principle across this literature is the exploitation of low-level statistical inconsistencies---noise patterns, compression artifacts, or boundary discontinuities---that betray the presence of a manipulation even when it is visually convincing~\cite{kwon_learning_2022,chen_enhancing_2025,zhu_mesoscopic_2024}.

ManTra-Net~\cite{wu_mantra-net_2019} operationalizes this principle through self-supervised training: a manipulation-trace feature extractor is trained to classify 385 distinct manipulation types, and forgery localization is then cast as a local anomaly detection problem using a ConvLSTM-based module that computes Z-score deviations across multiple window scales. While powerful in its generality, this approach does not explicitly model compression artifacts, which limits its sensitivity to the JPEG-specific traces prevalent in document images. CAT-Net~\cite{kwon_learning_2022} directly addresses this limitation by processing DCT coefficients through a dedicated stream that learns the statistical distribution of double-quantization residuals, combined with a standard RGB stream; the method relies on the observation that a region compressed twice leaves characteristic periodic patterns in the DCT coefficient histograms, whose structure depends on the ratio between the two quantization steps. PSCC-Net~\cite{liu_pscc-net_2022} pursues a different strategy, focusing on multi-scale spatial consistency rather than compression traces: a top-down path built on a lightweight HRNet backbone extracts multi-scale features, while a bottom-up path progressively refines manipulation masks from coarse to fine scales through a Spatio-Channel Correlation Module (SCCM) that captures both spatial and channel-wise correlations.

TruFor~\cite{guillaro_trufor_2023} extends this line of work by introducing a multimodal fusion framework that combines the RGB image with Noiseprint++, a learned noise-sensitive fingerprint trained via self-supervised contrastive learning on real images subjected to 512 distinct editing histories; both modalities are fused through a SegFormer-based transformer encoder, and the framework additionally produces a reliability map identifying regions where localization predictions are uncertain, along with a global integrity score. This reliability mechanism is particularly relevant in operational settings where false alarms carry a significant cost~\cite{guillaro_trufor_2023}. Building on the limitations of purely microscopic approaches, Mesorch~\cite{zhu_mesoscopic_2024} argues that manipulation operates simultaneously at the microscopic level through low-level forensic traces and at the macroscopic level through semantic object-level alterations, and addresses both by running a CNN branch and a Transformer branch in parallel: the CNN processes high-frequency DCT-enhanced features to capture local textural anomalies, while the Transformer processes low-frequency DCT-enhanced features to model global semantic context, with an adaptive weighting module dynamically adjusting the contribution of each scale.

These models have become standard references in image manipulation localization~\cite{du_forensichub_2025,ma_imdl-benco_2024}. However, they are designed and evaluated under assumptions---rich visual content, diverse textures, standard compression profiles---that do not hold for administrative document images, motivating the development of document-specific approaches.

\subsubsection{Specific Case of Documents.}
\label{sec:doc_localization}

Document images differ fundamentally from natural images in their forensic properties. The presence of text, regular layouts, and uniform backgrounds suppresses the boundary artifacts that natural image methods rely upon, while artifacts introduced by scanning, printing, or document conversion processes create additional confounding signals~\cite{dong_robust_2024,qu_towards_2023,luo_toward_2025}. Even a single substituted character can substantially alter the semantic content of a document~\cite{qu_towards_2023,dong_robust_2024,luo_toward_2025}, which distinguishes document forensics from natural image forensics where manipulations typically affect larger semantic regions. Accordingly, document-focused approaches have predominantly targeted pixel-level localization rather than whole-image detection~\cite{du_forensichub_2025,ma_imdl-benco_2024,qu_towards_2023}.

\begin{figure}[ht]
    \centering
    \begin{minipage}{0.45\linewidth}
        \begin{subfigure}{\linewidth}
            \centering
            \includegraphics[width=0.7\linewidth]{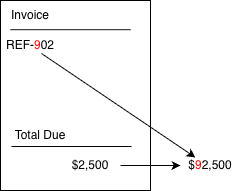}
            \caption{Copy-move}
            \vspace{0.4cm}
            \label{fig:copymove}
        \end{subfigure}
        \begin{subfigure}{\linewidth}
            \centering
            \includegraphics[width=\linewidth]{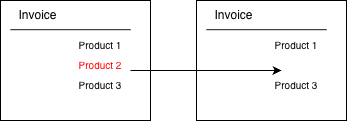}
            \caption{Coverage}
            \vspace{0.4cm}
            \label{fig:coverage}
        \end{subfigure}
        \begin{subfigure}{\linewidth}
            \centering
            \includegraphics[width=\linewidth]{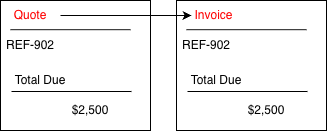}
            \caption{Generation}
            \label{fig:generation}
        \end{subfigure}
    \end{minipage}
    \hfill
    \begin{minipage}{0.4\linewidth}
        \begin{subfigure}{\linewidth}
            \centering
            \includegraphics[width=\linewidth]{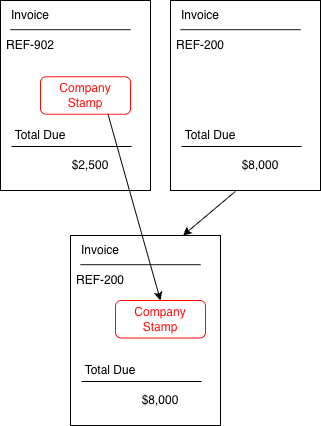}
            \caption{Splicing}
            \label{fig:splicing}
        \end{subfigure}
    \end{minipage}
    \caption{Illustration of the 4 manipulation types considered in document forensics~-~\textcolor{red}{Red} elements denote the tampered regions. \textbf{(a)}~Copy-move: a region is duplicated within the same document. \textbf{(b)}~Coverage: a patch conceals existing content. \textbf{(c)}~Generation: new pixel content is synthesized to replace a region. \textbf{(d)}~Splicing: a region extracted from a different source document is inserted into the target.}
    \label{fig:manipulation_types}
\end{figure}

As illustrated in Figure~\ref{fig:manipulation_types}, the manipulation operations encountered in document forensics---copy-move, splicing, generation, and coverage---are predominantly applied at the character or word level~\cite{qu_towards_2023,luo_toward_2025,artaud_find_2018}. These operations are typically confined to small areas and forensic traces they leave are subtle and easily masked by the uniform backgrounds features~\cite{chen_enhancing_2025,dong_robust_2024}.

DTD~\cite{qu_towards_2023} establishes the foundational framework for this domain by combining a Visual Perception Head with a Frequency Perception Head operating on raw DCT coefficients, fused through a Swin-Transformer encoder and a Multi-view Iterative Decoder (MID) that progressively aggregates features at multiple scales; to improve robustness against JPEG compression, the authors further propose Curriculum Learning for Tampering Detection (CLTD), a training paradigm that progressively increases compression difficulty with quality factors drawn from $[75, 100]$. FFDN~\cite{chen_enhancing_2025} extends this dual-branch design by identifying two residual weaknesses: the incomplete integration of frequency information into the RGB feature space, and the loss of fine-grained high-frequency traces during downsampling. It addresses these through a Visual Enhancement Module (VEM) that injects frequency information into the RGB stream using zero-initialized convolutions, and a Wavelet-like Frequency Enhancement (WFE) module that explicitly decomposes features into high- and low-frequency components. Both DTD and FFDN share a common architectural property that is central to this study: \textbf{they explicitly ingest the JPEG quantization table as an input to the network}, making their learned representations inherently dependent on the distribution of quantization tables seen during training.

In contrast, CAFTB~\cite{song_cross-attention_2025} operates in the noise domain: two parallel branches---one extracting spatial features from the RGB image, the other applying an SRM filter to expose global noise inconsistencies---are fused through a cross-attention module to integrate local and global forgery cues. TIFDM~\cite{dong_robust_2024} similarly enhances forgery traces from multiple domains through a dedicated deep module before feeding them into an encoder-decoder network with a multiscale attention module, with a particular emphasis on robustness to real-world distortions. Neither CAFTB nor TIFDM explicitly processes quantization tables, relying instead on implicit pixel-domain or noise-domain artifacts.

Recent benchmarking efforts have begun to unify the evaluation of these heterogeneous approaches. ForensicHub~\cite{du_forensichub_2025} provides a modular and configuration-driven architecture that decomposes forensic pipelines into interchangeable components across datasets, transforms, models, and evaluators, enabling cross-domain comparisons across the four main forensic tasks including document manipulation localization. Similarly, IMDL-BenCo~\cite{ma_imdl-benco_2024} standardizes the training and evaluation protocols for image manipulation detection and localization. Both frameworks highlight a persistent gap: the robustness of document forensic models to compression variability remains insufficiently characterized, as evaluations are typically restricted to standard quality factor ranges.

\subsection{JPEG Quantization as an Alteration Detector}
\label{sec:quantification}

JPEG compression artifacts have long constituted a central forensic signal in image authenticity analysis~\cite{wan_overview_2023,kwon_learning_2022,neal_krawetz_pictures_2007}. One of the earliest practical methods exploiting these artifacts is Error Level Analysis (ELA)~\cite{neal_krawetz_pictures_2007}, which detects inconsistencies in residual compression error across image regions: when a tampered image is re-saved at a fixed quality, authentic regions and manipulated regions exhibit different error levels because they carry distinct compression histories. Since these artifacts originate from block-based DCT coding followed by coefficient quantization, forensic cues are commonly analyzed through compression inconsistencies observed in the DCT domain~\cite{kwon_learning_2022,qu_towards_2023,chen_enhancing_2025}, whose manifestation is governed by the underlying quantization tables~\cite{wan_overview_2023,kwon_learning_2022}.

\begin{figure}[ht]
    \centering
    \includegraphics[width=\linewidth]{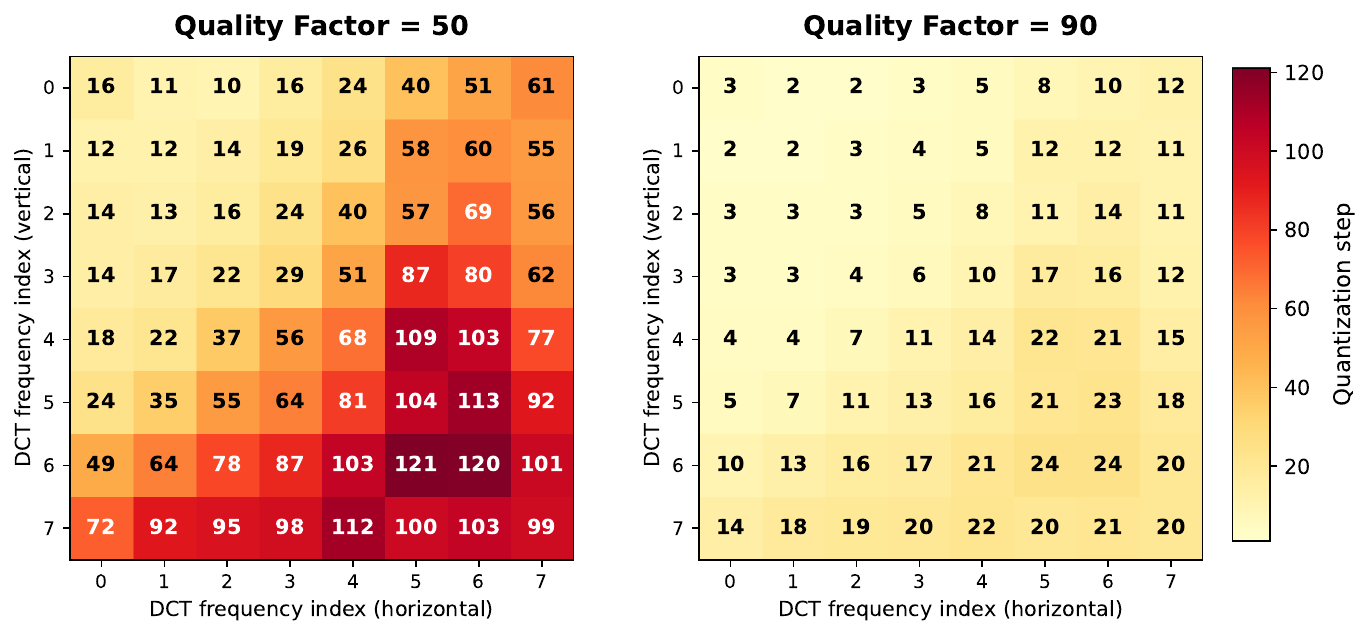}
    \caption{\textbf{Standard \textit{libjpeg} luminance quantization tables} at two quality factors. Each cell value represents the quantization step for the corresponding DCT frequency coefficient. At $q=50$ (left), high-frequency coefficients are aggressively quantized (steps up to 121), discarding fine detail. At $q=90$ (right), steps are much smaller (maximum 24), preserving more information. Both tables are derived from the same base matrix scaled by $q$~\cite{wan_overview_2023}, and represent only two points in the discrete family of tables reachable through standard quality factors.}
    \label{fig:qt_heatmap}
\end{figure}

JPEG compression partitions the image into non-overlapping $8 \times 8$ blocks, transforms each block into the frequency domain via the Discrete Cosine Transform (DCT), and then divides the resulting 64 coefficients component-wise by a $8 \times 8$ quantization matrix before rounding to the nearest integer~\cite{wan_overview_2023,kwon_learning_2022}. This quantization step is the primary source of information loss: large matrix entries aggressively suppress high-frequency coefficients, while small entries preserve detail at the cost of larger file size~\cite{wan_overview_2023,neal_krawetz_pictures_2007}. As illustrated in Figure~\ref{fig:qt_heatmap}, the conventional quality factor $q \in [1, 100]$ scales a reference matrix according to a fixed formula~\cite{wan_overview_2023}, producing quantization tables with markedly different coefficient magnitudes---and thus different compression artifacts---across quality levels. When a JPEG image undergoes a second compression cycle, the DCT coefficient histograms exhibit characteristic periodic patterns whose structure depends on the ratio between the two quantization matrices~\cite{wan_overview_2023,kwon_learning_2022}. It is precisely these double-quantization residuals that methods such as CAT-Net~\cite{kwon_learning_2022} and the Frequency Perception Head of DTD and FFDN~\cite{qu_towards_2023,chen_enhancing_2025} are designed to detect, which explains why the identity of the quantization matrix---rather than merely the quality level---is critical to their correct operation. The JPEG standard defines separate quantization tables for luminance (Y) and chrominance (Cb, Cr) channels~\cite{wan_overview_2023}; since the forensic models considered in this study operate exclusively on the luminance table~\cite{qu_towards_2023,chen_enhancing_2025,kwon_learning_2022}, all subsequent references to \textit{quantization table} denote the luminance table.

A key aspect of the JPEG standard is that quantization tables are not universally fixed: virtually all graphical applications and digital cameras rely on hard-coded, manufacturer-specific tables optimized for their own hardware and color pipelines~\cite{neal_krawetz_pictures_2007}. As a consequence, a given nominal quality level does not correspond to a unique quantization table across encoders. For example, a JPEG saved at quality 80\% in Adobe Photoshop uses quantization tables equivalent to quality 91\% in libjpeg~\cite{neal_krawetz_pictures_2007}. In practice, however, many existing forensic works implicitly assume that compression follows standard libjpeg-based encoding conventions: they parameterize compression exclusively through integer quality factors, which map to quantization tables via a fixed formula~\cite{wan_overview_2023} and thus cover only a small, discrete subset of the tables encountered in practice.

Among models that incorporate explicit compression augmentation, the quality factor ranges remain narrow: CAT-Net~\cite{kwon_learning_2022} uses factors in $[60, 100]$, DTD~\cite{qu_towards_2023} and FFDN~\cite{chen_enhancing_2025} apply random recompression restricted to $[75, 100]$, and TruFor~\cite{guillaro_trufor_2023}, the broadest of these, uses $[30, 100]$. Several other models, including ManTra-Net~\cite{wu_mantra-net_2019}, PSCC-Net~\cite{liu_pscc-net_2022}, and Mesorch~\cite{zhu_mesoscopic_2024}, include no explicit JPEG augmentation at all. All these configurations are derived from integer quality factors, which correspond to a small discrete subset of possible quantization tables and implicitly assume that compression follows standard tool-based encoding conventions~\cite{wan_overview_2023}. \textbf{Despite the known diversity of quantization tables in real-world imaging pipelines~\cite{neal_krawetz_pictures_2007}, no existing work has systematically analyzed the impact of this diversity on the performance of document manipulation localization models}, nor proposed training strategies to mitigate the resulting distribution shift.

\subsection{Proposed Methodology}
\label{sec:proposed_methodology}

To address the gap identified in the preceding sections, we propose:
\begin{enumerate}[noitemsep, topsep=2pt]
    \item A \textbf{dataset analysis} comparing the JPEG quantization table distributions across public scientific benchmarks and an operational insurance corpus, quantifying the extent of the heterogeneity gap (Section~\ref{sec:dataset_analysis}).
    \item An \textbf{experimental protocol} designed to isolate the impact of JPEG quantization table distribution shift on forgery localization performance, through a factorial design that crosses two recompression pipelines with two structurally contrasting models (Section~\ref{sec:benchmark_protocol}).
\end{enumerate}

%% file: sections/donnees.tex
\section{Dataset Analysis}
\label{sec:dataset_analysis}

\subsection{Datasets}
\label{sec:datasets}

The datasets considered in this study were selected to cover the majority of publicly available document image datasets containing both altered and non-altered samples, spanning a wide range of document types, manipulation methods, and acquisition conditions. This selection includes DocTamper~\cite{qu_towards_2023}, RTM~\cite{luo_toward_2025}, T-SROIE~\cite{wang_tampered_2022}, SROIE~\cite{huang_icdar2019_2019}, FUNSD~\cite{jaume_funsd_2019}, and Find-it / Find-it Again~\cite{artaud_find_2018,martinez_tornes_jeu_2023} as public benchmarks, alongside the MAIF operational image corpus used to evaluate model behavior on real operational data exhibiting heterogeneous acquisition and compression pipelines. Due to privacy and confidentiality constraints, the underlying MAIF document images cannot be publicly released. From this image corpus, however, we derive \textsc{DocQT}, a public dataset of header-extracted luminance quantization tables used to define the \textit{Real-QT} condition; these tables do not contain document content and are released, together with the material required to reproduce the compression protocol, at \url{https://github.com/Kyliroco/Improving-Document-Forgery-Localization-Robustness-via-Diverse-JPEG-Quantization-Tables}. This combination enables a comprehensive evaluation of model robustness across both controlled and operational environments.

We establish a terminological distinction between \textit{document images} and \textit{natural images}. Document images refer to administrative or insurance-related documents processed in operational contexts. In such images, character-level tampering can be visually imperceptible: the absence of texture suppresses the boundary artifacts that detection methods rely upon, making localization significantly more challenging~\cite{chen_enhancing_2025,dong_robust_2024,luo_toward_2025,qu_towards_2023}. Documents processed in operational contexts exhibit substantial diversity both in content type---ranging from administrative forms to photographic evidence captured on smartphones~\cite{artaud_find_2018,qu_towards_2023,luo_toward_2025}---and in image format, as they are commonly submitted as native PDFs, scanned PDFs, or raster images (JPEG or PNG)~\cite{nayef_smartdoc-qa_2015}. Raster photographs are predominantly stored in JPEG format, which introduces compression artifacts whose characteristics depend directly on the quantization table used during encoding~\cite{neal_krawetz_pictures_2007,kwon_learning_2022}.

\begin{table}[ht]
  \centering
  \caption{Datasets considered in this work. CM: copy-move; SP: splicing;
    GN: generation; IP: inpainting; CV: coverage. The symbol $\star$ denotes
    unaltered reference sets used for false positive evaluation only.}
  \label{tab:datasets}
  \resizebox{\textwidth}{!}{
    \begin{tabular}{llcccc}
      \hline
      \textbf{Dataset}                                                          & \textbf{Subset}     & \textbf{\# Images} & \textbf{Mod.\ Type}
                                                                                & \textbf{\# Altered} & \textbf{\# QT}                                                        \\
      \hline
      \multirow{5}{*}{DocTamper~\cite{qu_towards_2023}}
                                                                                & Train               & 120\,000           & CM / GN / SP           & 120\,000 & 1            \\
                                                                                & Test                & 30\,000            & CM / GN / SP           & 30\,000  & 1            \\
                                                                                & FCD                 & 2\,000             & CM / GN / SP           & 2\,000   & 1            \\
                                                                                & SCD                 & 18\,000            & CM / GN / SP           & 18\,000  & 1            \\
      \hline
      \multirow{2}{*}{RTM~\cite{luo_toward_2025}}
                                                                                & Train               & 5\,803             & CM / GN / SP / IP / CV & 4\,000   & 1            \\
                                                                                & Test                & 3\,197             & CM / GN / SP / IP / CV & 2\,000   & 1            \\
      \hline
      \multirow{2}{*}{T-SROIE~\cite{wang_tampered_2022}}
                                                                                & Train               & 626                & GN                     & 626      & 1            \\
                                                                                & Test                & 360                & GN                     & 360      & 1            \\
      \hline
      SROIE$^\star$~\cite{huang_icdar2019_2019}                                 & --                  & 973                & --                     & --       & 6            \\
      \hline
      Find-it~\cite{artaud_find_2018}                                           & --                  & 1\,180             & CM / GN / SP / IP      & 240      & 6            \\
      \hline
      FUNSD$^\star$~\cite{jaume_funsd_2019}                                     & --                  & 199                & --                     & --       & --           \\
      \hline
      Find-it Again~\cite{martinez_tornes_jeu_2023}                             & --                  & 988                & CM / GN / SP           & 163      & --           \\
      \hline
      \makecell[l]{MAIF operational corpus$^\star$ \\ (real documents, France)} & --                  & 13\,455            & --                     & --       & \textbf{859} \\
      \hline
    \end{tabular}
  }
\end{table}

\subsection{Quantization Table Diversity Across Corpora}
\label{sec:qt_diversity}

A central motivation of this study is the discrepancy between the JPEG compression diversity observed in public benchmarks and that encountered in operational settings. As reported in Table~\ref{tab:datasets}, SROIE~\cite{huang_icdar2019_2019} and Find-it~\cite{artaud_find_2018} are the only public datasets exhibiting more than a single luminance quantization table, with 6 distinct configurations each arising from the heterogeneity of scanning equipment; all other public datasets---DocTamper~\cite{qu_towards_2023}, RTM~\cite{luo_toward_2025}, and T-SROIE~\cite{wang_tampered_2022}---expose a single uniform quantization table throughout, consistent with systematic generation using a standard encoder at a fixed quality factor.

\begin{figure}[ht]
  \centering
  \includegraphics[width=\linewidth]{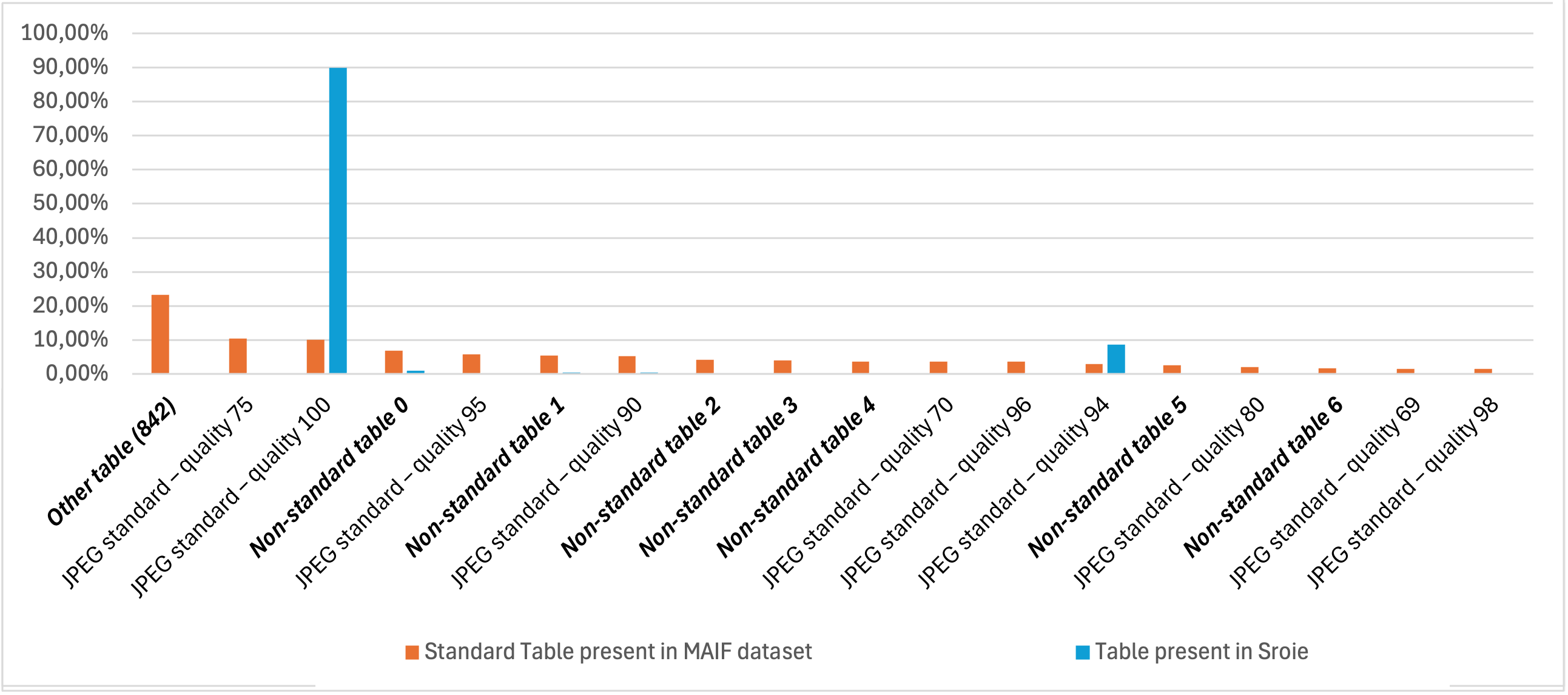}
  \caption{Distribution of the most frequent luminance JPEG quantization tables in the \textsc{MAIF} operational dataset (orange) and SROIE~\cite{huang_icdar2019_2019} (blue). Each bar represents a distinct quantization table, ranked by frequency of occurrence in the operational corpus. The first bar aggregates the 851 remaining tables, each individually representing less than 25\% of the operational dataset. While a single table accounts for over 90\% of SROIE images, no dominant configuration emerges in the operational corpus, confirming its substantially higher compression diversity.}
  \label{fig:jpeg_pareto_comparison}
\end{figure}

In contrast, the MAIF dataset exhibits 859 distinct luminance quantization tables, reflecting the diversity of acquisition devices, scanning software, and document conversion chains encountered in insurance workflows~\cite{dong_robust_2024,nayef_smartdoc-qa_2015}. As illustrated in Figure~\ref{fig:jpeg_pareto_comparison}, the distribution follows a pronounced long-tail pattern: even SROIE, the most diverse public benchmark, concentrates the vast majority of its images within 6 configurations, whereas the operational corpus distributes images across hundreds of non-standard tables with no dominant configuration. Although this frequency profile comes from a single French insurance workflow, it aggregates customer-submitted documents acquired through diverse devices and software chains, so overlap in quantization-table support with other administrative settings is plausible even if mixture weights differ. These observations confirm that public benchmarks do not capture the quantization table variability present in operational document environments~\cite{dong_robust_2024,luo_toward_2025,artaud_find_2018}, and directly motivate the \textit{Real-QT} experimental condition described in Section~\ref{sec:recompression_pipelines}.

%% file: sections/benchmark_protocol.tex
\section{Benchmark Protocol}
\label{sec:benchmark_protocol}

This section describes the experimental protocol designed to isolate the effect of JPEG quantization table distribution shift on document manipulation localization. Two training pipelines are tested: \textit{Standard-QT}, which applies classical quality factor augmentation, and \textit{Real-QT}, which uses quantization tables extracted from real operational document headers. These pipelines are crossed with two structurally contrasting models to form a factorial design.

\subsection{JPEG Recompression Pipelines}
\label{sec:recompression_pipelines}

The central experimental variable of this study is the JPEG quantization table distribution encountered during both training and evaluation. Two recompression pipelines are defined and applied independently, yielding a factorial design that combines training pipeline with evaluation recompression condition.

The first pipeline, referred to as \textit{Standard-QT}, recompresses all images with quality factors drawn uniformly at random from $[30, 100]$ using OpenCV, covering the broadest quality factor range reported in the literature (see Section~\ref{sec:quantification} for a detailed comparison across methods).

The second pipeline, referred to as \textit{Real-QT}, recompresses all images using quantization tables sampled from \textsc{DocQT}, a dataset extracted from the JPEG headers of the MAIF operational image corpus following the fingerprinting approach described in~\cite{neal_krawetz_pictures_2007}. This dataset yields over 800 distinct luminance quantization tables reflecting heterogeneous real-world acquisition and processing pipelines.
% Since OpenCV does not support arbitrary quantization matrices, this recompression is performed using Pillow.

Each model variant is evaluated under three conditions: no forced recompression (\textit{Orig.}), \textit{Standard-QT} recompression (\textit{Std}), and \textit{Real-QT} recompression (\textit{Real}). This factorial design enables us to disentangle the effect of the training distribution from that of the evaluation distribution on localization and false positive performance.

\subsection{Training Pipeline}
\label{sec:training_pipeline}

All images are normalized using ImageNet statistics to match the expected input distribution of the pretrained backbones used by both models. Data augmentation is applied during training only: horizontal and vertical flips with probability 0.5, random rotation by multiples of 90 degrees with probability 0.5, random brightness and contrast adjustment, and Gaussian blur with probability 0.2. These transformations are chosen to improve generalization without degrading the low-level forensic artifacts that both models rely upon~\cite{wu_mantra-net_2019,kwon_learning_2022}.

\subsection{Methods Used}
\label{sec:methods_used}

All experiments are conducted within the ForensicHub framework~\cite{du_forensichub_2025}, which provides a unified training and evaluation infrastructure across forensic tasks and natively implements both FFDN~\cite{chen_enhancing_2025} and Mesorch~\cite{zhu_mesoscopic_2024}. These two models are selected because they represent opposite ends of a structural spectrum directly relevant to the central question of this paper---the impact of JPEG quantization table distribution shift on localization performance.

As described in Section~\ref{sec:doc_localization}, FFDN inherits from DTD~\cite{qu_towards_2023} a Frequency Perception Head that ingests the quantization table as an explicit input, making it by design sensitive to any mismatch between training-time and inference-time quantization tables. Mesorch, as described in Section~\ref{sec:natural_localization}, does not exploit the quantization table explicitly: it relies on implicit pixel-domain and DCT-domain artifacts captured through parallel CNN and Transformer branches, which makes it a particularly informative counterpoint for assessing the role of explicit quantization table awareness under distribution shift.

The remaining models available in ForensicHub---such as ManTra-Net~\cite{wu_mantra-net_2019}, CAT-Net~\cite{kwon_learning_2022}, or PSCC-Net~\cite{liu_pscc-net_2022}---are not retained in the main comparison because they do not sharpen the same explicit-versus-implicit quantization conditioning contrast. Preliminary CAT-Net runs under the same protocol followed the same qualitative direction as FFDN under \textit{Real-QT}, but with lower absolute performance and weaker cross-dataset generalization, so the main analysis remains centered on the two more discriminative endpoints of this spectrum.

Two model variants are trained, one under each pipeline: \textbf{FFDN-\textit{Std}} and \textbf{Mesorch-\textit{Std}} are trained under \textit{Standard-QT}; \textbf{FFDN-\textit{Real}} and \textbf{Mesorch-\textit{Real}} are trained under \textit{Real-QT}.

\subsection{Data}
\label{sec:data}

\paragraph{Dataset Splits.}
Training data is drawn from four sources: the full DocTamper training set~\cite{qu_towards_2023} (120{,}000 images), and 5{,}000 crops each from the training portions of Find-it~\cite{artaud_find_2018}, Find-it Again~\cite{martinez_tornes_jeu_2023}, and T-SROIE~\cite{wang_tampered_2022}. Additionally, 400 crops from FUNSD~\cite{jaume_funsd_2019}---approximately half the dataset---and a subset of unaltered documents from the MAIF operational image corpus are included as negative examples to expose the models to authentic document statistics during training. RTM~\cite{luo_toward_2025} is deliberately excluded from training in order to evaluate the capacity of models trained exclusively on synthetically generated manipulations to generalize to real manual forgeries. For datasets providing official splits, only the official training partition is used; for datasets without predefined splits, the training subset is defined prior to any preprocessing to prevent data leakage. All remaining data are reserved for evaluation.

\subsection{Evaluation Criteria}
\label{sec:evaluation}

All models produce a pixel-level probability map $P \in [0,1]^{H \times W}$, from which a binary prediction mask $\hat{M}$ is derived by thresholding at $\tau = 0.5$:

\begin{equation}
  \hat{M}(i,j) = \mathbb{I}\bigl(P(i,j) \geq \tau\bigr).
\end{equation}

Given the ground-truth binary mask $M_{gt}$ provided by each dataset, the following pixel-level quantities are defined:

\begin{align}
  TP_{\text{pix}} & = \textstyle\sum_{i,j} \mathbb{I}\bigl(\hat{M}(i,j)=1 \wedge M_{gt}(i,j)=1\bigr), \\
  FP_{\text{pix}} & = \textstyle\sum_{i,j} \mathbb{I}\bigl(\hat{M}(i,j)=1 \wedge M_{gt}(i,j)=0\bigr), \\
  FN_{\text{pix}} & = \textstyle\sum_{i,j} \mathbb{I}\bigl(\hat{M}(i,j)=0 \wedge M_{gt}(i,j)=1\bigr).
\end{align}

\subsubsection{Localization on Tampered Images.}
On images containing at least one altered pixel, localization performance is jointly measured by the pixel-level \textbf{F1 score} and \textbf{Intersection over Union}~(IoU)~\cite{ma_imdl-benco_2024,du_forensichub_2025}:

\begin{equation}
  F_1^{\text{pix}} = \frac{2\,TP_{\text{pix}}}{2\,TP_{\text{pix}} + FP_{\text{pix}} + FN_{\text{pix}}}, \qquad
  \mathrm{IoU} = \frac{TP_{\text{pix}}}{TP_{\text{pix}} + FP_{\text{pix}} + FN_{\text{pix}}}.
\end{equation}

Both metrics are computed per image and averaged across the evaluation set.

\subsubsection{False Positive Rate on Unaltered Images.}
For unaltered reference sets, the ground-truth mask is identically zero ($M_{gt} = \mathbf{0}$), so every predicted positive pixel is by definition a false positive. The \textbf{pixel-level false positive rate} is defined as:

\begin{equation}
  \mathrm{FPR}_{\text{pix}} = \frac{FP_{\text{pix}}}{H \times W}, \label{fpr_pix}
\end{equation}

where $H \times W$ is the total number of pixels. This quantity is computed per image and reported as a mean across the reference set~\cite{guillaro_trufor_2023}. A low $\mathrm{FPR}_{\text{pix}}$ combined with high F1 and IoU on tampered images constitutes the primary robustness criterion of this study: it captures whether a model can accurately localize tampered regions without being misled by unfamiliar JPEG compression profiles into flagging authentic documents.

%% file: sections/Results.tex
\section{Results}
\label{sec:results}

% \subsection{Results}
% \label{sec:results_tables}

% Tables~\ref{tab:f1_results},~\ref{tab:iou_results} and~\ref{tab:fpr_results} report respectively pixel-level F1, IoU, and mean $\mathrm{FPR}_{\text{pix}}$ for each combination of training pipeline and evaluation recompression condition (\textit{Orig.}, \textit{Std}, \textit{Real}).
\paragraph{Quantization Table Distribution as a Primary Source of Generalization Failure.}
Across all DocTamper subsets, evaluation without recompression (\textit{Orig.}) consistently yields the highest scores, with substantial drops under both recompression conditions. The benefit of operational training is, however, strongly conditioned on the model's capacity to exploit this distributional alignment explicitly. On DT-Test under \textit{Real-QT} evaluation, \textbf{FFDN-Real gains 14.5 F1 points over FFDN-Std} (0.853 vs.\ 0.708), with consistent improvements across FCD and SCD. By contrast, Mesorch-\textit{Real} and Mesorch-\textit{Std} achieve nearly identical scores under the same condition (0.704 vs.\ 0.703). This contrast directly reflects the architectural distinction identified in Section~\ref{sec:doc_localization}: FFDN's Frequency Perception Head can exploit the distributional match through its explicit quantization table input, whereas Mesorch has no mechanism to bind its representations to a specific quantization matrix~\cite{chen_enhancing_2025,zhu_mesoscopic_2024}. \textbf{Architectures that explicitly condition on the quantization table therefore provide a meaningful robustness advantage in operational compression pipelines}~\cite{kwon_learning_2022,wan_overview_2023}.

% === TABLE F1 ===
\begin{table*}[ht]
  \centering
  \caption{Pixel-level F1 score on tampered evaluation sets. Columns correspond to model
    variants trained under \textit{Standard-QT} (suffix \textit{-Std}) or \textit{Real-QT}
    (suffix \textit{-Real}). Rows report three evaluation conditions per dataset subset:
    \textit{Orig.}\ (no recompression), \textit{Std}\ (\textit{Standard-QT}
    recompression), \textit{Real}\ (\textit{Real-QT} recompression). \textbf{Bold}:
    best result per row within each architecture family (FFDN and Mesorch separately).}
  \label{tab:f1_results}
  \resizebox{\textwidth}{!}{
    \begin{tabular}{llc|cc|cc}
      \hline
                                                               &                       &                &
      \textbf{FFDN-\textit{Std}}~\cite{chen_enhancing_2025}    &
      \textbf{FFDN-\textit{Real}}~\cite{chen_enhancing_2025}   &
      \textbf{Mesorch-\textit{Std}}~\cite{zhu_mesoscopic_2024} &
      \textbf{Mesorch-\textit{Real}}~\cite{zhu_mesoscopic_2024}                                                                                                             \\
      \hline
      \multirow{9}{*}{DocTamper~\cite{qu_towards_2023}}
                                                               & \multirow{3}{*}{Test}
                                                               & \textit{Orig.}        & 0.927          & \textbf{0.954} & 0.751          & \textbf{0.818}                  \\
                                                               &                       & \textit{Std}   & 0.633          & \textbf{0.647} & \textbf{0.676} & 0.578          \\
                                                               &                       & \textit{Real}  & 0.708          & \textbf{0.853} & 0.703          & \textbf{0.704} \\
      \cmidrule(lr){2-7}
                                                               & \multirow{3}{*}{FCD}
                                                               & \textit{Orig.}        & 0.942          & \textbf{0.951} & \textbf{0.555} & 0.509                           \\
                                                               &                       & \textit{Std}   & 0.536          & \textbf{0.537} & \textbf{0.556} & 0.476          \\
                                                               &                       & \textit{Real}  & 0.630          & \textbf{0.832} & \textbf{0.540} & 0.530          \\
      \cmidrule(lr){2-7}
                                                               & \multirow{3}{*}{SCD}
                                                               & \textit{Orig.}        & 0.879          & \textbf{0.900} & 0.591          & \textbf{0.709}                  \\
                                                               &                       & \textit{Std}   & 0.516          & \textbf{0.542} & \textbf{0.514} & 0.472          \\
                                                               &                       & \textit{Real}  & 0.604          & \textbf{0.776} & 0.537          & \textbf{0.587} \\
      \hline
      \multirow{3}{*}{RTM~\cite{luo_toward_2025}}
                                                               & \multirow{3}{*}{--}
                                                               & \textit{Orig.}        & \textbf{0.048} & 0.040          & 0.058          & \textbf{0.074}                  \\
                                                               &                       & \textit{Std}   & \textbf{0.034} & 0.031          & \textbf{0.048} & 0.042          \\
                                                               &                       & \textit{Real}  & \textbf{0.030} & 0.023          & 0.051          & \textbf{0.057} \\
      \hline
      \multirow{3}{*}{T-SROIE~\cite{wang_tampered_2022}}
                                                               & \multirow{3}{*}{--}
                                                               & \textit{Orig.}        & \textbf{0.900} & 0.893          & \textbf{0.873} & 0.865                           \\
                                                               &                       & \textit{Std}   & 0.746          & \textbf{0.798} & \textbf{0.833} & 0.754          \\
                                                               &                       & \textit{Real}  & 0.727          & \textbf{0.827} & \textbf{0.838} & 0.798          \\
      \hline
      \multirow{3}{*}{Find-it~\cite{artaud_find_2018}}
                                                               & \multirow{3}{*}{--}
                                                               & \textit{Orig.}        & 0.296          & \textbf{0.418} & 0.309          & \textbf{0.364}                  \\
                                                               &                       & \textit{Std}   & 0.168          & \textbf{0.216} & \textbf{0.223} & 0.189          \\
                                                               &                       & \textit{Real}  & 0.186          & \textbf{0.303} & 0.252          & \textbf{0.253} \\
      \hline
      \multirow{6}{*}{Find-it Again~\cite{martinez_tornes_jeu_2023}}
                                                               & \multirow{3}{*}{Test}
                                                               & \textit{Orig.}        & 0.130          & \textbf{0.158} & 0.068          & \textbf{0.212}                  \\
                                                               &                       & \textit{Std}   & 0.052          & \textbf{0.058} & 0.059          & \textbf{0.077} \\
                                                               &                       & \textit{Real}  & \textbf{0.073} & 0.067          & 0.075          & \textbf{0.125} \\
      \cmidrule(lr){2-7}
                                                               & \multirow{3}{*}{Val}
                                                               & \textit{Orig.}        & \textbf{0.204} & 0.201          & 0.063          & \textbf{0.259}                  \\
                                                               &                       & \textit{Std}   & 0.071          & \textbf{0.078} & \textbf{0.069} & 0.064          \\
                                                               &                       & \textit{Real}  & \textbf{0.114} & 0.076          & 0.043          & \textbf{0.183} \\
      \hline
    \end{tabular}
  }
\end{table*}
% === TABLE FPR ===
\begin{table*}[ht]
  \centering
  \caption{Mean pixel-level false positive rate ($\mathrm{FPR}_{\mathrm{pix}}$, lower is
    better) on unaltered reference sets. Same column and row structure as
    Table~\ref{tab:f1_results}. \textbf{Bold}: best (lowest) result per row within each
    architecture family.}
  \label{tab:fpr_results}
  \resizebox{\textwidth}{!}{
    \begin{tabular}{lc|cc|cc}
      \hline
                                                               &                &
      \textbf{FFDN-\textit{Std}}~\cite{chen_enhancing_2025}    &
      \textbf{FFDN-\textit{Real}}~\cite{chen_enhancing_2025}   &
      \textbf{Mesorch-\textit{Std}}~\cite{zhu_mesoscopic_2024} &
      \textbf{Mesorch-\textit{Real}}~\cite{zhu_mesoscopic_2024}                                                                                                                                             \\
      \hline
      \multirow{3}{*}{SROIE~\cite{huang_icdar2019_2019}}
                                                               & \textit{Orig.} & $3.50\times10^{-5}$          & $\mathbf{2.82\times10^{-5}}$ & $3.21\times10^{-5}$          & $\mathbf{1.60\times10^{-5}}$ \\
                                                               & \textit{Std}   & $\mathbf{2.77\times10^{-5}}$ & $2.36\times10^{-5}$          & $4.12\times10^{-5}$          & $\mathbf{2.88\times10^{-5}}$ \\
                                                               & \textit{Real}  & $4.65\times10^{-5}$          & $\mathbf{1.69\times10^{-5}}$ & $4.06\times10^{-5}$          & $\mathbf{4.00\times10^{-5}}$ \\
      \hline
      \multirow{3}{*}{FUNSD~\cite{jaume_funsd_2019}}
                                                               & \textit{Orig.} & $2.49\times10^{-4}$          & $\mathbf{6.01\times10^{-5}}$ & $1.48\times10^{-4}$          & $\mathbf{5.98\times10^{-5}}$ \\
                                                               & \textit{Std}   & $\mathbf{7.63\times10^{-5}}$ & $1.42\times10^{-4}$          & $\mathbf{1.76\times10^{-4}}$ & $1.78\times10^{-4}$          \\
                                                               & \textit{Real}  & $8.53\times10^{-4}$          & $\mathbf{7.79\times10^{-5}}$ & $1.87\times10^{-4}$          & $\mathbf{1.47\times10^{-4}}$ \\
      \hline
      \multirow{3}{*}{\textsc{MAIF}}
                                                               & \textit{Orig.} & $5.64\times10^{-4}$          & $\mathbf{1.46\times10^{-4}}$ & $2.06\times10^{-4}$          & $\mathbf{9.66\times10^{-5}}$ \\
                                                               & \textit{Std}   & $2.39\times10^{-4}$          & $\mathbf{1.30\times10^{-4}}$ & $2.03\times10^{-4}$          & $\mathbf{1.84\times10^{-4}}$ \\
                                                               & \textit{Real}  & $2.32\times10^{-4}$          & $\mathbf{8.69\times10^{-5}}$ & $2.37\times10^{-4}$          & $\mathbf{1.84\times10^{-4}}$ \\
      \hline
    \end{tabular}
  }
\end{table*}
% === TABLE IoU ===
\begin{table*}[ht]
  \centering
  \caption{Pixel-level IoU score on tampered evaluation sets. \textbf{Bold}: best result per row within each
    architecture family.}
  \label{tab:iou_results}
  \resizebox{\textwidth}{!}{
    \begin{tabular}{llc|cc|cc}
      \hline
                                                               &                       &                &
      \textbf{FFDN-\textit{Std}}~\cite{chen_enhancing_2025}    &
      \textbf{FFDN-\textit{Real}}~\cite{chen_enhancing_2025}   &
      \textbf{Mesorch-\textit{Std}}~\cite{zhu_mesoscopic_2024} &
      \textbf{Mesorch-\textit{Real}}~\cite{zhu_mesoscopic_2024}                                                                                                             \\
      \hline
      \multirow{9}{*}{DocTamper~\cite{qu_towards_2023}}
                                                               & \multirow{3}{*}{Test}
                                                               & \textit{Orig.}        & 0.882          & \textbf{0.923} & 0.692          & \textbf{0.762}                  \\
                                                               &                       & \textit{Std}   & 0.579          & \textbf{0.597} & \textbf{0.617} & 0.525          \\
                                                               &                       & \textit{Real}  & 0.653          & \textbf{0.809} & 0.644          & \textbf{0.647} \\
      \cmidrule(lr){2-7}
                                                               & \multirow{3}{*}{FCD}
                                                               & \textit{Orig.}        & 0.895          & \textbf{0.913} & \textbf{0.505} & 0.469                           \\
                                                               &                       & \textit{Std}   & 0.484          & \textbf{0.489} & \textbf{0.505} & 0.432          \\
                                                               &                       & \textit{Real}  & 0.574          & \textbf{0.776} & \textbf{0.488} & 0.480          \\
      \cmidrule(lr){2-7}
                                                               & \multirow{3}{*}{SCD}
                                                               & \textit{Orig.}        & 0.806          & \textbf{0.837} & 0.498          & \textbf{0.618}                  \\
                                                               &                       & \textit{Std}   & 0.439          & \textbf{0.466} & \textbf{0.430} & 0.402          \\
                                                               &                       & \textit{Real}  & 0.524          & \textbf{0.699} & 0.450          & \textbf{0.505} \\
      \hline
      \multirow{3}{*}{RTM~\cite{luo_toward_2025}}
                                                               & \multirow{3}{*}{--}
                                                               & \textit{Orig.}        & \textbf{0.039} & 0.032          & 0.046          & \textbf{0.061}                  \\
                                                               &                       & \textit{Std}   & \textbf{0.026} & 0.024          & \textbf{0.037} & 0.033          \\
                                                               &                       & \textit{Real}  & \textbf{0.023} & 0.018          & 0.040          & \textbf{0.046} \\
      \hline
      \multirow{3}{*}{T-SROIE~\cite{wang_tampered_2022}}
                                                               & \multirow{3}{*}{--}
                                                               & \textit{Orig.}        & 0.766          & \textbf{0.827} & \textbf{0.803} & 0.799                           \\
                                                               &                       & \textit{Std}   & 0.663          & \textbf{0.720} & \textbf{0.755} & 0.677          \\
                                                               &                       & \textit{Real}  & 0.643          & \textbf{0.753} & \textbf{0.761} & 0.724          \\
      \hline
      \multirow{3}{*}{Find-it~\cite{artaud_find_2018}}
                                                               & \multirow{3}{*}{--}
                                                               & \textit{Orig.}        & 0.257          & \textbf{0.370} & 0.275          & \textbf{0.321}                  \\
                                                               &                       & \textit{Std}   & 0.147          & \textbf{0.192} & \textbf{0.196} & 0.167          \\
                                                               &                       & \textit{Real}  & 0.165          & \textbf{0.269} & 0.223          & \textbf{0.223} \\
      \hline
      \multirow{6}{*}{Find-it Again~\cite{martinez_tornes_jeu_2023}}
                                                               & \multirow{3}{*}{Test}
                                                               & \textit{Orig.}        & 0.096          & \textbf{0.123} & 0.051          & \textbf{0.158}                  \\
                                                               &                       & \textit{Std}   & 0.036          & \textbf{0.045} & 0.044          & \textbf{0.057} \\
                                                               &                       & \textit{Real}  & \textbf{0.053} & 0.048          & 0.057          & \textbf{0.092} \\
      \cmidrule(lr){2-7}
                                                               & \multirow{3}{*}{Val}
                                                               & \textit{Orig.}        & \textbf{0.153} & 0.152          & 0.044          & \textbf{0.191}                  \\
                                                               &                       & \textit{Std}   & 0.051          & \textbf{0.057} & \textbf{0.049} & 0.045          \\
                                                               &                       & \textit{Real}  & \textbf{0.083} & 0.053          & 0.030          & \textbf{0.136} \\
      \hline
    \end{tabular}
  }
\end{table*}

\paragraph{Standard Quality Factor Augmentation Is Not a Sufficient Proxy for Operational Conditions.}
\textit{Real-QT} evaluation consistently yields higher localization scores than \textit{Standard-QT}, challenging the assumption---common across the models reviewed in Section~\ref{sec:quantification}---that broad quality factor augmentation adequately proxies real-world compression variability. Two factors explain this. First, quality factor augmentation covers only the discrete subset of libjpeg-compatible matrices, missing the application-specific tables that dominate the operational distribution (Section~\ref{sec:qt_diversity}). Second, the low-quality end of the $[30, 100]$ range is unrepresentative of operational documents, which concentrate around moderate quality values and therefore preserve more forensic signal than the \textit{Standard-QT} distribution implies~\cite{dong_robust_2024}.

\paragraph{The Synthetic-to-Real Gap Remains an Open Problem.}
On RTM~\cite{luo_toward_2025}, excluded from training, all variants achieve near-zero performance (peak F1: 0.074 for Mesorch-\textit{Real}), revealing a limitation orthogonal to the compression distribution question investigated here. The forensic traces exploited by models trained on synthetic manipulations --- double-quantization residuals and boundary discontinuities --- are largely absent in images tampered manually by professional editors~\cite{kwon_learning_2022,wu_mantra-net_2019}. Addressing this gap requires either large-scale manual tampering data or architectures reasoning on semantic inconsistencies independently of low-level artifacts~\cite{zhu_mesoscopic_2024,guillaro_trufor_2023}.

\paragraph{Benchmark Diversity Does Not Reflect Operational Diversity.}
Models achieve substantially higher scores on T-SROIE (F1 up to 0.900) than on Find-it (best F1: 0.418) and Find-it Again (best F1: 0.259), despite equal training exposure. T-SROIE's uniform JPEG configuration aligns with the DocTamper training signal, whereas the multi-tool structure of Find-it and the lossless format of Find-it Again expose failure modes that single-configuration benchmarks obscure~\cite{du_forensichub_2025,ma_imdl-benco_2024}. This observation reinforces the quantization table diversity analysis in Section~\ref{sec:qt_diversity}: benchmarks relying on a single quantization table provide an overly optimistic estimate of model performance in real-world deployment scenarios.

\paragraph{False Positive Rates.}
On the operational corpus, \textbf{FFDN-Real under Real-QT achieves a $\mathrm{FPR}_{\mathrm{pix}}$ of $8.69 \times 10^{-5}$}, nearly an order of magnitude lower than FFDN-\textit{Std} at \textit{Orig.} ($5.64\times10^{-4}$). Mesorch-\textit{Real} attains its lowest FPR without recompression ($9.66\times10^{-5}$), while recompression increases it, peaking on FUNSD under \textit{Real-QT} ($1.87\times10^{-3}$). Across all conditions, FPR remains at most in the $10^{-3}$ range, confirming that both models produce acceptably low false positive rates on unaltered documents. Critically, the reduction achieved by \textit{Real-QT} training validates the hypothesis of Section~\ref{sec:problem_statement}: \textbf{quantization table mismatch significantly contributes to false alarms in deployment}.

\subsection{Limitations and Perspectives}
\label{sec:limitations}

The comparison is restricted to two architectures selected for structural contrast (Section~\ref{sec:methods_used}). Preliminary CAT-Net runs under the same protocol followed the same qualitative trend as FFDN under \textit{Real-QT}, but with lower absolute performance and weaker cross-dataset generalization, so broader architectural validation remains needed~\cite{kwon_learning_2022}. The operational table distribution derives from a single insurance corpus which, despite 859 distinct configurations, may not represent industries. Finally, the threshold $\tau = 0.5$ may be suboptimal in deployment scenarios with asymmetric false positive / false negative costs~\cite{guillaro_trufor_2023,ma_imdl-benco_2024}.

Future work should extend this analysis to architectures with intermediate frequency conditioning, test chrominance-table and chrominance-DCT conditioning alongside luminance cues, construct benchmarks from pristine PNG documents with controlled recompression and manually edited documents~\cite{du_forensichub_2025,ma_imdl-benco_2024}, and explore domain adaptation techniques that do not require access to operational documents at training time.

%% file: sections/Conclusion.tex
\section{Conclusion}
\label{sec:conclusion}

This paper investigated the impact of JPEG quantization table distribution shifts on document manipulation localization models deployed in operational insurance workflows. Through a controlled factorial study comparing FFDN~\cite{chen_enhancing_2025} and Mesorch~\cite{zhu_mesoscopic_2024}, selected for contrasting quantization table awareness (Section~\ref{sec:methods_used}), we showed that training under operationally calibrated quantization tables yields substantial localization gains---up to 14.5 F1 points on DocTamper~\cite{qu_towards_2023}---but only for architectures explicitly ingesting the quantization table as input. On the false positive side, this protocol reduces $\mathrm{FPR}_{\text{pix}}$ on the MAIF operational image corpus by nearly an order of magnitude ($8.69 \times 10^{-5}$ vs.\ $5.64 \times 10^{-4}$). These findings show that standard quality factor augmentation does not adequately proxy operational compression diversity, and that architectures conditioning on quantization tables provide a meaningful robustness advantage for real-world deployment. Additionally, near-zero performance on RTM~\cite{luo_toward_2025} confirms that the synthetic-to-real manipulation gap remains an open problem orthogonal to compression, motivating future benchmarks built from pristine documents, controlled recompression, and manually edited forgeries.